\documentclass[conference]{IEEEtran}
\IEEEoverridecommandlockouts

\usepackage{cite}
\usepackage{float}
\usepackage{graphicx}
\usepackage{amssymb}

\usepackage{amsmath}
\usepackage{mathtools}
\usepackage{mathrsfs}
\usepackage{svg}

\usepackage{algorithm}
\usepackage{algpseudocode}

\usepackage{nccmath}

\usepackage{amsthm}
\usepackage[usenames,dvipsnames,svgnames,table]{xcolor}
\theoremstyle{plain}

\newtheorem{definition}{Definition}

\newtheorem{problem}{Problem}
\newtheorem{mission}{Mission}

\usepackage{float}
\floatstyle{plaintop}
\restylefloat{table}
\usepackage{MnSymbol}%
\usepackage{wasysym}%
\usepackage{caption}
\usepackage{subcaption}
\usepackage{graphicx}
\usepackage{import}
\usepackage{lipsum}
\usepackage{soul}

\graphicspath{ {./images/} }

\usepackage[colorinlistoftodos]{todonotes}
 \renewcommand{\baselinestretch}{0.9045}

\newcommand{\diegoqb}[2]{%
    \todo[size=\footnotesize, color=blue!30]{#1}%
    \st{#2}%
}

\usepackage[hidelinks]{hyperref}%

\begin{document}

\title{
Cooperative Aerial Transportation of a Payload with a Sliding Cable Attachment
\thanks{
Acknowledgement Statement - Research reported in this publication was supported in part by funding
provided by the National Aeronautics and Space Administration (NASA), under award number
80NSSC25M7087, Michigan Space Grant Consortium (MSGC).

The authors are with the Distributed Multi-Agent Robotic Systems ($\Delta$MARS Lab), Department of Electrical and Computer Engineering, Oakland University, Rochester, MI 48309, USA. 
}
}

\author{\IEEEauthorblockN{Sebastian Tilaguy Lezama}
\IEEEauthorblockA{\textit{dept. Electrical and Computer Engineering} \\
\textit{Oakland University}\\
Rochester, MI, US \\
\texttt{tilaguylez@oakland.edu}
\and
\IEEEauthorblockN{Diego S. D'Antonio}
\IEEEauthorblockA{\textit{dept. Electrical and Computer Engineering} \\
\textit{Oakland University}\\
Rochester, MI, US \\
\texttt{dantonio@oakland.edu}
}}}

\maketitle

\begin{abstract}
This paper presents a dual-quadrotor hook-suspended transport system for cooperative aerial manipulation of a payload that is not rigidly attached to a fixed point along the cable. Two quadrotors are connected by a single inextensible cable that passes through a frictionless hook mounted on the payload, allowing the cable to slide between its two segments. This introduces a cable-distribution variable that couples the quadrotor motion, cable geometry, and payload dynamics.  A feasible reference-generation framework is formulated using the desired payload pose, cable distribution, and system shape while satisfying the cable-length constraint and elliptical workspace. The quadrotors track the geometrically generated trajectories while maintaining positive cable tension and respecting actuator limits.  The payload pose is estimated from the system geometry, and its measured position is used only to validate the desired trajectory. The approach is evaluated through experimental demonstrations.
\end{abstract}

\section{INTRODUCTION}





Cooperative aerial systems have expanded the capabilities of unmanned aerial vehicles (UAVs) beyond surveillance and conventional logistics. Through distributed actuation and coordinated motion, multiple UAVs can perform tasks that exceed the physical limitations of a single platform, including heavy-payload transportation, infrastructure interaction, and emergency response
\cite{Kardasz_2016,Garg_2023,Ostermann_2025}.
Aerial logistics has therefore received increasing attention for rapid operation in remote or hard-to-reach environments, motivating scalable and autonomous multi-robot transportation strategies
\cite{Moshref-Javadi_2021}.

Cable-suspended transportation provides several advantages for aerial manipulation. By separating the payload from the aerial platform, the cable reduces the direct transmission of disturbances and allows the UAVs to maintain a safer distance from the transported object. It can also facilitate payload attachment and release while the aerial vehicles remain in flight
\cite{Eskandaripour_2023,Pounds_2011}.
Early studies primarily considered payloads suspended from a single UAV, which inherently limits the available lifting capacity and controllability of the payload. Cooperative multi-UAV systems address these limitations by distributing the payload weight among multiple aerial vehicles and coordinating the cable forces applied to the load
\cite{Michael_2011}.

\begin{figure}[t]
    \centering
    \includegraphics[width=0.6\linewidth]{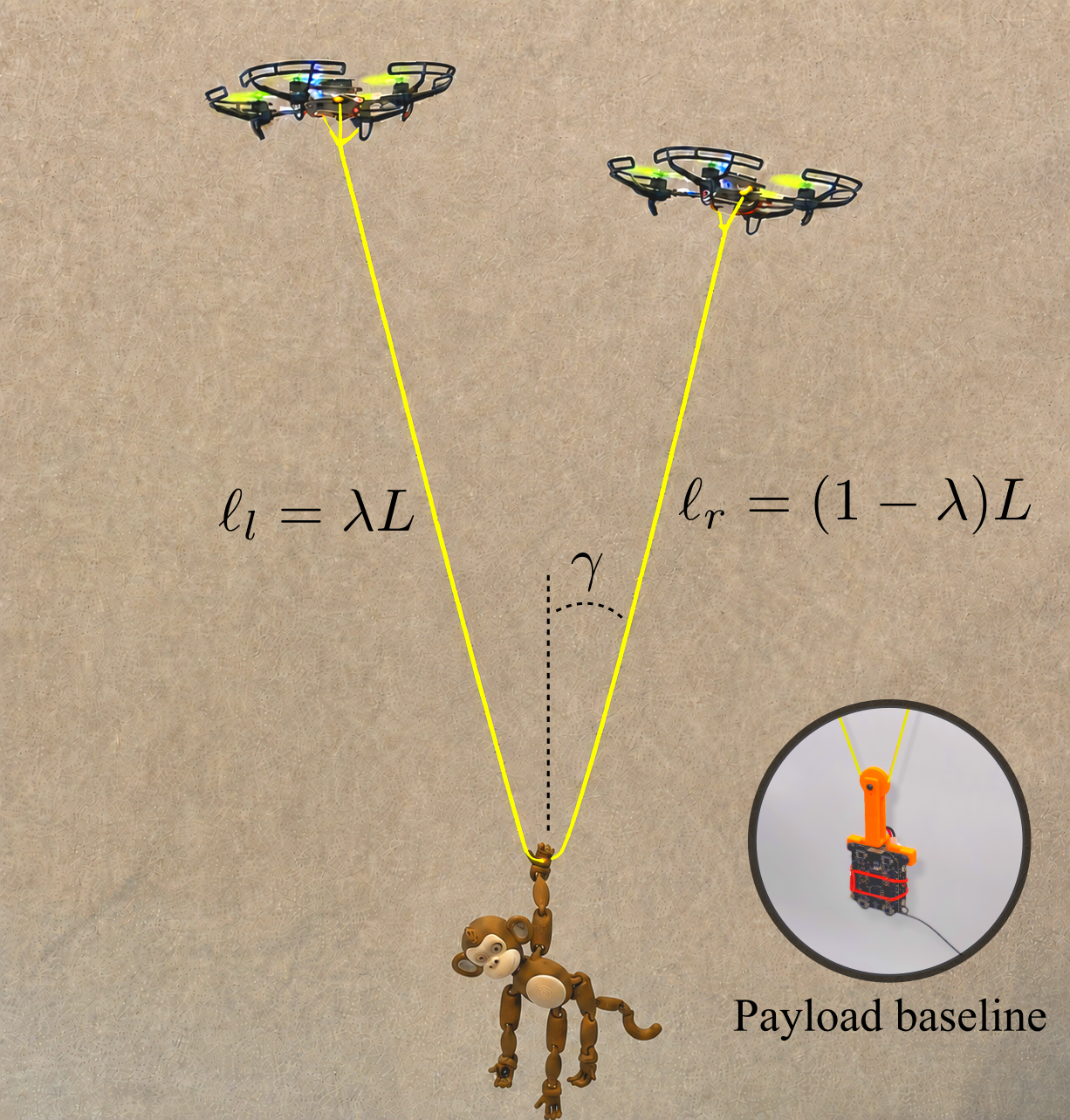}
    \caption{
    DQ-HSTS: A single cable connects the two quadrotors and passes through a hook rigidly attached to the payload, allowing the cable to slide.
    }
    \vspace{-2em}
    \label{fig:mainfigure}
\end{figure}


Several approaches have addressed cooperative cable-suspended transportation through differential flatness, coordinated rotorcraft motion, inertial sensing, autonomous hook-based grasping, and nonlinear model predictive control \cite{Kotaru2017DifferentialFlatness,GarciaDeMarina2019Flexible,Xie2020CooperativeTransport,Antal2023HookBasedGrasping,Sun2023NonlinearMPC}. These methods enable trajectory generation, cooperative transport, payload grasping, and constrained full-pose manipulation. However, they generally assume fixed cable attachment points and do not consider cable-length redistribution through a sliding payload-mounted hook.

Cable-suspended aerial lifting is commonly divided into \textit{setup}, \textit{pull}, and \textit{raise} stages \cite{Cruz_2015,Estevez_2024}. While the pull and raise stages have received significant attention, setup often requires manual cable attachment, reducing autonomy and limiting operation in hazardous or inaccessible environments. A \textit{catenary robot}, formed by two aerial vehicles connected by a flexible cable, can manipulate the cable to interact with objects \cite{DAntonio_2021}. However, engaging a payload-mounted hook allows the cable to slide, coupling UAV motion, cable geometry, payload dynamics, and cable-length distribution.
Motivated by this gap, this work introduces the \textit{Dual-Quadrotor Hook-Suspended Transport System} (DQ-HSTS), shown in Fig.~\ref{fig:mainfigure}. Two quadrotors are connected by a single cable passing through a frictionless hook rigidly attached to the payload. Unlike conventional systems, the cable slides through the hook, redistributing its total length between the two cable segments.

This paper addresses quadrotor coordination for payload-pose and cable-distribution control without fixed cable attachments. The contributions are threefold:
\textit{i)} the DQ-HSTS, in which a cable slides through a payload-mounted hook;
\textit{ii)} trajectory generation parameterized by payload pose, cable distribution $\lambda$, and system shape $\gamma$; and
\textit{iii)} coordinated trajectory tracking under tension, geometric, and actuator constraints, validated in experiments.

\begin{figure}
    \centering
    \includegraphics[width=0.55\linewidth]{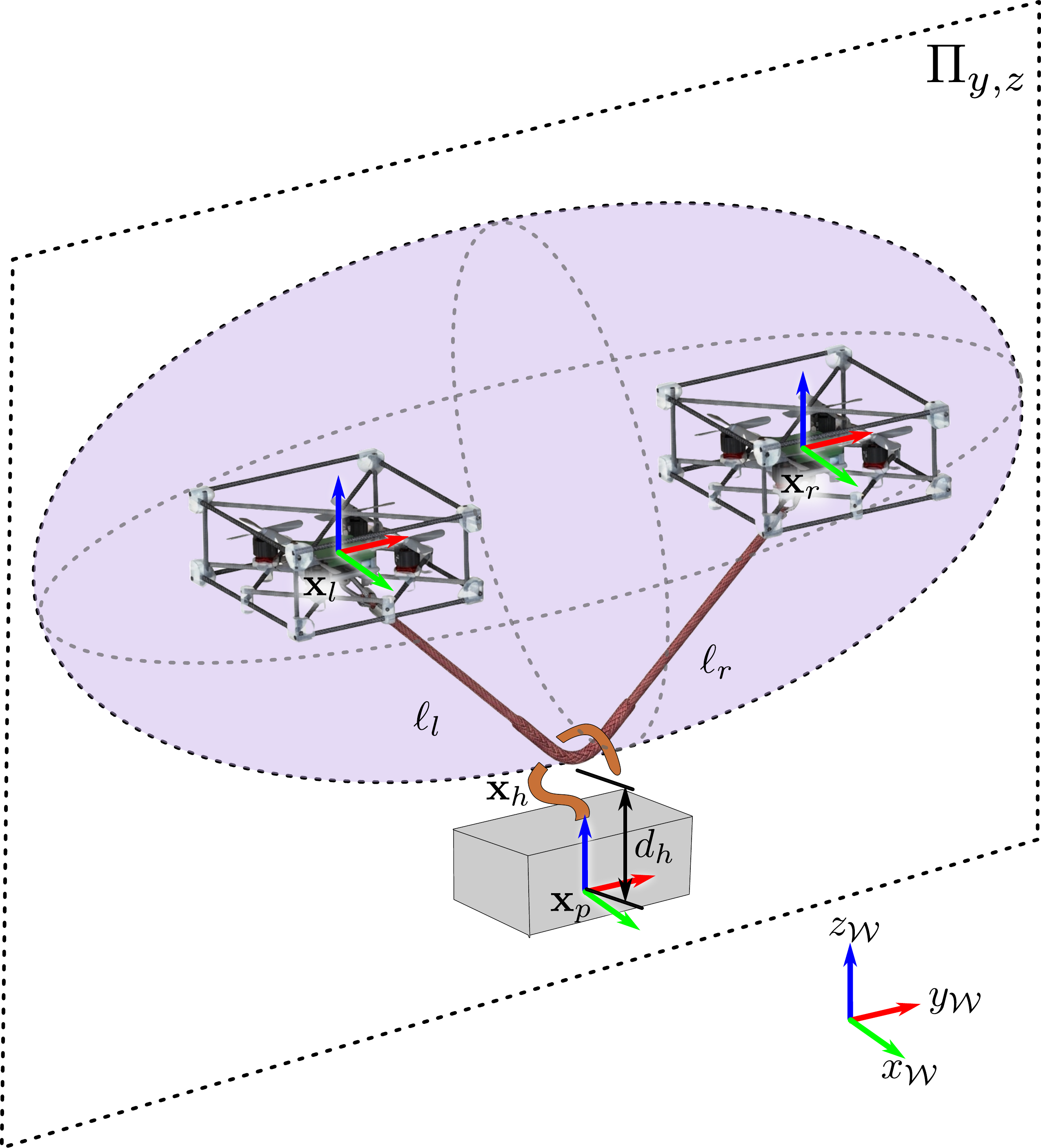}
    \caption{Free body diagram for a DQ-HSTS.}
    \label{fig:free-body}
    \vspace{-1em}
\end{figure}



\section{Sliding hook-payload}

\label{sec:problem_statement}

Cooperative aerial transportation using cables traditionally assumes that the payload is rigidly attached to one or more aerial vehicles through fixed cable connections. In contrast, this work considers a \emph{Dual-Quadrotor Hook-Suspended Transport System} (DQ-HSTS), where two quadrotors are connected by a single cable that passes through a frictionless hook rigidly attached to the payload (see Fig.~\ref{fig:free-body}). Unlike conventional systems, the cable is allowed to slide freely through the hook. This dynamically changes the distribution of the cable length between the left and right segments while preserving the total cable length, heavily coupling the cable geometry, payload dynamics, and quadrotor motion.

\subsection{System Description and Kinematics}

\begin{definition}[Dual-Quadrotor Hook-Suspended Transport System]
The DQ-HSTS is a mechanical system composed of two quadrotors attached to the ends of a single ideal cable. The cable is assumed to be massless, inextensible, perfectly flexible, and taut. It passes through a frictionless hook, rigidly fixed at a constant distance from the payload, allowing the cable to slide while maintaining equal tension in both segments.
\end{definition}

Let $\mathcal{W}$ denote the inertial frame with its $z$-axis pointing upward. The positions of the left and right quadrotors are denoted by $\mathbf{x}_l, \mathbf{x}_r \in \mathbb{R}^3$, respectively. The payload's center of mass (COM) is located at $\mathbf{x}_p \in \mathbb{R}^3$, and its attitude is represented by the rotation matrix $R_p \in SO(3)$.

Although the dynamics are governed by full three-dimensional Newton--Euler equations, this work analyzes a planar configuration to study the fundamental coupling of the sliding hook. The system evolves in the vertical $\Pi_{y,z}$ plane, and the rotational motion of both the quadrotors and the payload is restricted to pitch rotations about the out-of-plane $x$-axis, denoted by the angle $\phi_p$.

The hook is rigidly attached to the payload at a constant distance $d_h$ above its COM. Therefore, its position is given by
$
\mathbf{x}_h
=
\mathbf{x}_p
+
d_h R_{\phi_p} \mathbf{e}_z,
$
where $\mathbf{e}_z = \begin{bmatrix} 0 & 0 & 1 \end{bmatrix}^{\top}$ is the unit vector along the $z$-axis. 


\begin{figure*}[t]
    \centering
    \includegraphics[width=0.8\linewidth]{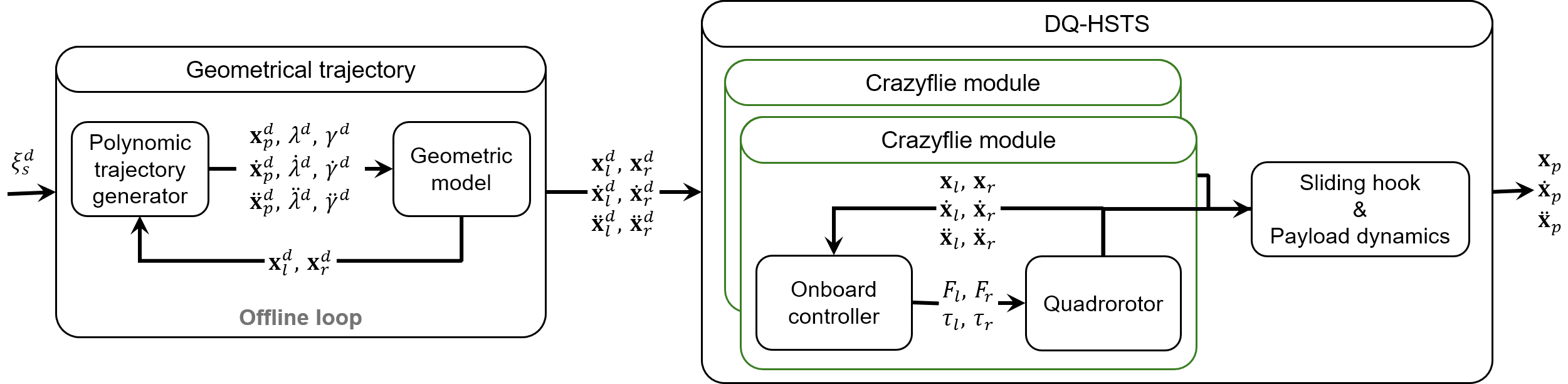}
    \caption{Control diagram of DQ-HSTS for payload positioning.}
    \label{fig:ctrol_diag}
    \vspace{-1.5em}
\end{figure*}

The hook divides the cable into two segments of lengths $\ell_l$ and $\ell_r$. Because the cable is inextensible, the total length constraint is
$
\ell_l + \ell_r = L,
\label{eq:length_constraint}
$
where $L$ is the constant total cable length. The cable distribution is parameterized by a scalar $\lambda \in (0,1)$, which specifies the fraction of the total cable length allocated to the left segment. Accordingly, $\ell_l = \lambda L$ and $\ell_r = (1-\lambda)L$.

For fixed quadrotor positions, the cable-length constraint dictates that the hook must satisfy:
\begin{equation}
\left\| \mathbf{x}_l - \mathbf{x}_h \right\|
+
\left\| \mathbf{x}_r - \mathbf{x}_h \right\|
=
L.
\label{eq:ellipse}
\end{equation}
This geometrically confines the hook to the outline of an ellipse (or prolate spheroid in 3D) whose foci coincide with the current positions of the two quadrotors. To ensure the payload remains suspended beneath the aerial vehicles, the physically admissible hook configuration is restricted to the lower half of this ellipse.

\subsection{System Dynamics}

Unlike conventional dual-cable transportation systems, the segment lengths $\ell_l$ and $\ell_r$ are not independently actuated; they evolve according to the passive sliding motion of the cable. Each quadrotor $i \in \{l,r\}$ is equipped with four propellers generating a total thrust $f_i$ and a body moment $\mathbf{M}_i$. The translational and rotational dynamics of the quadrotors are described by the Newton--Euler equations
\begin{align}
m_i\ddot{\mathbf{x}}_i &= -m_ig\mathbf{e}_z + f_i R_i \mathbf{e}_z + \mathbf{F}_{c,i}, \\
J_i\dot{\boldsymbol{\Omega}}_i + \boldsymbol{\Omega}_i \times J_i\boldsymbol{\Omega}_i &= \mathbf{M}_i,
\end{align}
where $m_i$ and $J_i$ are the mass and inertia matrix of quadrotor $i$, $R_i$ is its attitude, $\boldsymbol{\Omega}_i$ is its angular velocity, and $\mathbf{F}_{c,i}$ denotes the cable force acting on the quadrotor.

Under the frictionless hook assumption, both cable segments share an equal tension magnitude $T$. Defining the unit directional vectors from the hook to each quadrotor as
$$
\mathbf{e}_l = \frac{\mathbf{x}_l-\mathbf{x}_h}{\|\mathbf{x}_l-\mathbf{x}_h\|},
\qquad
\mathbf{e}_r = \frac{\mathbf{x}_r-\mathbf{x}_h}{\|\mathbf{x}_r-\mathbf{x}_h\|},
$$
the cable forces acting on the quadrotors are $\mathbf{F}_{c,l} = -T \mathbf{e}_l$ and $\mathbf{F}_{c,r} = -T \mathbf{e}_r$. The resultant lifting force applied at the hook is
\begin{equation}\label{eq:resultant-force-hook}
\mathbf{F}_h = T \left( \mathbf{e}_l + \mathbf{e}_r \right).
\end{equation}

The payload is modeled as a rigid body with mass $m_p$ and inertia $J_p$. Its dynamics are governed by
\begin{align}
m_p\ddot{\mathbf{x}}_p &= -m_pg\mathbf{e}_z + \mathbf{F}_h, \\
J_p\dot{\boldsymbol{\Omega}}_p + \boldsymbol{\Omega}_p \times J_p\boldsymbol{\Omega}_p &= \mathbf{r}_h \times \mathbf{F}_h,
\end{align}
where $\mathbf{r}_h = d_h R_{\phi_p} \mathbf{e}_z$ is the distance vector from the payload's COM to the hook.

\subsection{Problem Formulation}

Because the system configuration depends heavily on the unactuated sliding cable variable $\lambda$, recovering the complete state of the DQ-HSTS requires integrating both the geometric constraints and the system dynamics. The control objective is therefore bifurcated into reference generation and trajectory tracking.

\begin{problem}[Feasible Reference Generation]
\label{prob:reference_generation}
Given a desired payload-position trajectory $\mathbf{x}_p^d(t)$ and a fixed cable length $L$, determine geometrically feasible configuration references for the cable distribution $\lambda^d(t)$ and system shape parameter $\gamma^d(t)$. 
These parameters must generate corresponding hook $\mathbf{x}_h^d(t)$ and quadrotor trajectories $\mathbf{x}^{d}_l(t), \mathbf{x}^{d}_r(t)$ that satisfy the inextensible cable constraints while actively avoiding known obstacles in the workplace (e.g., narrow corridors).
\end{problem}

Once a physically admissible topology is generated, the quadrotors must be coordinated to generate the necessary cable tensions at the hook to manipulate the payload.

\begin{problem}[Payload Position and Cable-Distribution Control]
\label{prob:payload_pulley_control}
Given the feasible reference trajectories derived in Problem \ref{prob:reference_generation},
use the existing onboard controllers to track these references and thereby achieve indirect tracking of the desired payload position and cable distribution.
The approach assumes the cable remains taut ($T > 0$), preserving the hook configuration on the instantaneous lower-half ellipse, and sufficient actuation capacity for quadrotor reference tracking.
\end{problem}




\section{Trajectory generation and control}

In order to achieve cooperative transport involving a sliding hook, the architecture for indirect payload position control, based on quadrotor positioning, is divided into two cascaded stages (Fig. \ref{fig:ctrol_diag}) that directly address the formulated problems.

\subsection{Payload Reference Generation}
\label{subsec:hook_reference_generation}

The desired trajectory is formulated in the three-dimensional workspace through $N$ key points, collected in $\Xi^d \in \mathbb{R}^{N\times 5}$, where $\xi_s^d=[ \mathbf{x}_{p,s}^d, \lambda_s^d, \gamma_s^d ]^\top$ represents the initial point of the $s$-th sub-trajectory. $\mathbf{x}_{p,s}^d \in \mathbb{R}^3$ is the payload position vector, $\lambda\in(0,1)$, representing the cable length distribution, and $\gamma\in\left(0, \pi/2\right)$, representing the angle between each cable segment and the payload's vertical axis.

For payload movement\footnote{If the load position reference does not change and only $\lambda$ and/or $\gamma$ vary, $\Delta t$ is calculated in proportion to the total mission time: $\Delta t = T_T / (N-1)$.}, we defined $V_{\mathrm{max}}$ as a parameter to ensure very slow movement, thereby keeping the system close to the steady state. The total trajectory time $T_T$ and the time interval for each sub-trajectory $\Delta t_s$ are proportional to waypoint distances and are increased until the peak reference speed satisfies \(V_{\mathrm{eff}}\).

\begin{algorithm}[!b]
    \caption{Trajectory time allocation under velocity constraints}
    \label{alg:trajectory_time}
    \begin{algorithmic}[1]
        \Require $T_{\mathrm{req}}>0$, $V_{\max}>0$, $\delta_t\geq1$, $N$ payload waypoints $\mathbf{x}_{p,j}^d$
        \Ensure Segment durations $\Delta t_j$ and total time $T_T$
        
        \State $V_{\mathrm{eff}}\gets V_{\max}/\delta_t$
        \State $d_j\gets\|\mathbf{x}_{p,j+1}^d-\mathbf{x}_{p,j}^d\|$, $j=0,\ldots,N-2$  \Comment{Require $d_j>0$ for all $j$}
        \State $D\gets\sum d_j$
        \State $T_T\gets\max(T_{\mathrm{req}},D/V_{\mathrm{eff}})$
        \Repeat
            \State $\Delta t_j\gets T_Td_j/D$ for all $j$
            \State Generate the polynomial trajectory using $\Delta t_j$
            \State Compute $v_{\mathrm{peak}}$ over the trajectory
            \If{$v_{\mathrm{peak}}>V_{\mathrm{eff}}$}
                \State $T_T\gets T_Tv_{\mathrm{peak}}/V_{\mathrm{eff}}$
            \EndIf
        \Until{$v_{\mathrm{peak}}\leq V_{\mathrm{eff}}$}
        \State \Return $\Delta t_j,\,T_T$
    \end{algorithmic}
\end{algorithm}

Each time interval $\Delta t_s$ and its corresponding pair of desired points $(\xi^d_{s-1}, \xi^d_{s})$ are used to implement the minimum-snap algorithm proposed by \cite{Mellinger2011}. This algorithm yields a set of degree-$m$ polynomials as the time-indexed reference,

\begin{equation}
x_{p,s}(\tau)= c_{s,0}+...+c_{s,m}\tau^m =\sum_{k=0}^m {\mathbf{c}_{s,k}\tau^k},
\end{equation}
with $\mathbf{c}_{s,k}$ is the vector of coefficients of the $s$-th polynomial, $\tau=t-t_s$ is the relative time at each segment. Algorithm \ref{alg:trajectory_time} calculates the maximum velocity reached by the payload in each segment; if the established threshold is exceeded, the total trajectory time is rescaled and the polynomial is recalculated.

$$
v_{\mathrm{peak}} = \left \{ \max_t {\|\dot{\mathbf{x}}_{p}^d(t) \|} ~\colon~ v_{\mathrm{peak}} < V_{\mathrm{eff}} =V_{\mathrm{max}}/\delta_t \right\}
$$

$\lambda_s^d$ and $\gamma_s^d$ are subsequently modified by bounded quintic interpolations, depending on the robot's workspace. The polynomial coefficients are obtained minimizing the square of the fourth derivative.
$$
\min_c{
    J=\sum_s{
        \int_0^{\Delta t_s}{
            \left\| \frac{d^4x_{p,s}(\tau)}{d\tau^4} \right\|^2d\tau
            }
    }
},
\qquad 
A_{eq}\mathbf{c}=b_{eq}
$$
where the constraints impose exact passage through the waypoints, as well as continuity of velocity, acceleration, jerk, and snap between segments, and rest conditions at the trajectory's start and end points.

Once the optimal values of the coefficients $\mathbf{c}_{s,k}$ have been determined, the desired trajectory $\mathbf{x}_p^d(t)$ is found by evaluating each polynomial over its corresponding time interval.

\subsection{Vehicles Trajectory Transformation from the Payload}\label{subsec:geometrical_model}

Once the desired position of the payload is obtained, the necessary transformations are performed to map the reference to the position of the quadrotors. Given that, they are the only bodies in the system equipped with a control loop. The first transformation is to the desired hook position $\mathbf{x}_h^d$, which is obtained from the payload pose via the rigid attachment constraint:
\begin{equation}
\mathbf{x}_h^d(t) = \mathbf{x}_p^d(t) + d_h R_{\phi_p^d} \mathbf{e}_z.
\label{eq:desired_hook_position}
\end{equation}

Its velocity and acceleration are obtained by successive differentiation:
\begin{align}
\dot{\mathbf{x}}_h^d(t) &= \dot{\mathbf{x}}_p^d(t) + d_h \dot{R}_{\phi_p^d} \mathbf{e}_z, \label{eq:desired_hook_velocity} \\
\ddot{\mathbf{x}}_h^d(t) &= \ddot{\mathbf{x}}_p^d(t) + d_h \ddot{R}_{\phi_p^d} \mathbf{e}_z. \label{eq:desired_hook_acceleration}
\end{align}

For the trajectories considered here, the desired payload angle and its derivatives are set to zero ($R_{\phi_p}^d=I$), these expressions simplify to $\mathbf{x}_h^d = \mathbf{x}_p^d+d_h\mathbf{e}_z$ and $\dot{\mathbf{x}}_h^d = \dot{\mathbf{x}}_p^d$.

\begin{figure}[t]
    \centering
    \includegraphics[width=0.9\linewidth, height=4cm]{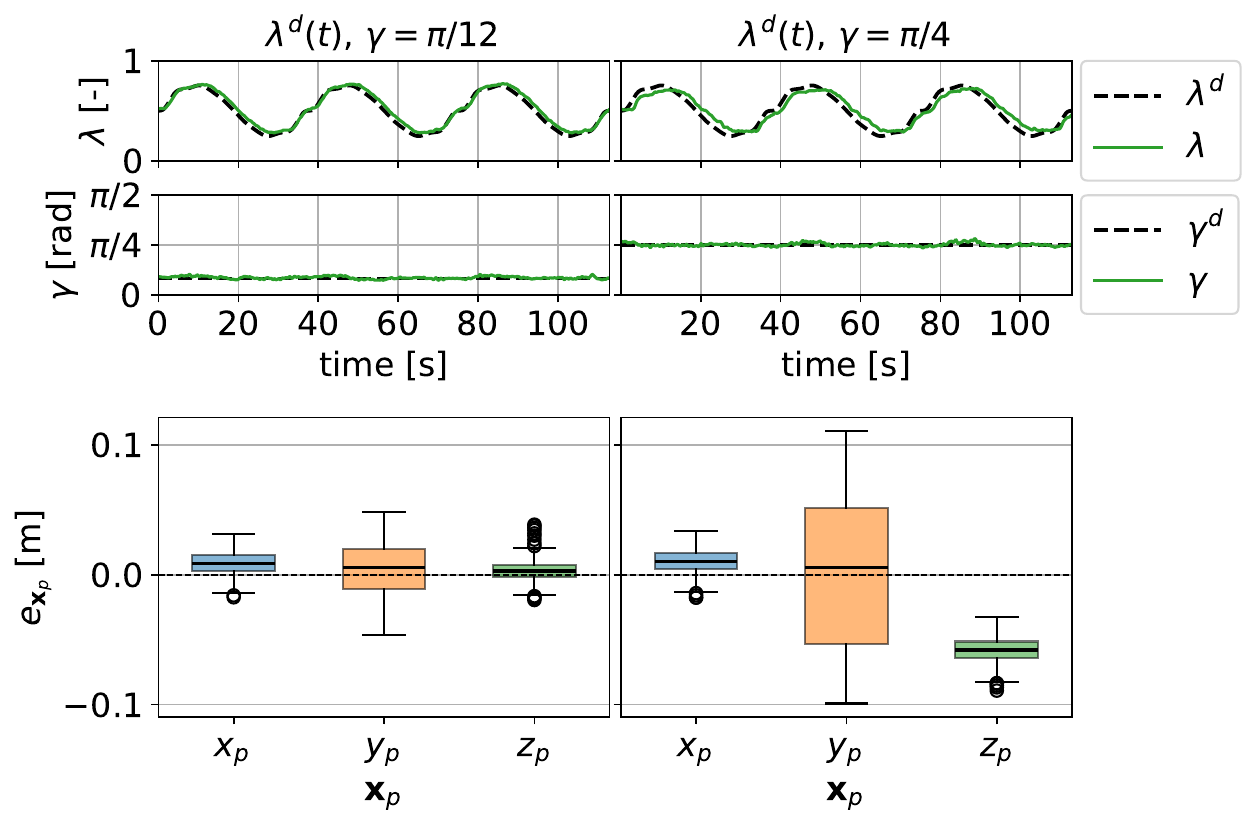}
    \caption{Experiment (i.a): Fixed payload position under $\lambda^d$ oscillations. Top: $\lambda^d(t)$ for two constant values of $\gamma^d$. Bottom: payload-position errors.}
    \label{fig:Ex1_lambda_swing}
    \vspace{-1em}
\end{figure}

To determine the desired quadrotor states $\mathbf{x}_l^{d}$ and $\mathbf{x}_r^{d}$, we used the \textit{prolate spheroid} geometric property of the sliding cable. Assuming the robot motion is around the equilibrium state, the unit direction vectors for the left and right cable segments are 
\begin{equation} \label{eq:planar-cable-dir}
    \mathbf{e}_l^d = \begin{bmatrix} 0 \\ -\sin{\gamma(t)} \\ \cos{\gamma(t)} \end{bmatrix}, \qquad
    \mathbf{e}_r^d = \begin{bmatrix} 0 \\ \sin{\gamma(t)} \\ \cos{\gamma(t)} \end{bmatrix}.
\end{equation}

The horizontal components of the resultant hook force naturally cancel out under this balanced assumption. The desired quadrotor positions are consequently defined by the hook position and the shape variables:
\begin{equation} \label{eq:drone-desired-position}
    \begin{aligned}
        \mathbf{x}^{d}_l(t) &= \mathbf{x}_h^d(t) + \lambda(t)L\mathbf{e}_l^d,\\
        \mathbf{x}_r^{d}(t) &= \mathbf{x}_h^d(t) + (1-\lambda(t))L\mathbf{e}_r^d.
    \end{aligned}
\end{equation}

The variables $\lambda$ and $\gamma$ command the spatial position of the DQ-HSTS. The focal distance (for $\lambda=0.5$, over the plane $\Pi_{y,z}$, Fig. \ref{fig:free-body}) is $L\sin{\gamma}$. A small $\gamma$ results in a narrow horizontal location but requires the quadrotors to fly higher, whereas a large $\gamma$ creates a wide, vertically compact system. The parameter $\lambda$ governs lateral asymmetry; if $\lambda \neq 0.5$, the payload shifts laterally relative to the midpoint of the quadrotors. The $i\in\{l,r\}$ vehicle velocity and acceleration references are
\begin{align}\label{eq:drone-desired-vel-acc}
    \dot{\mathbf{x}}^{d}_i(t) &= \dot{\mathbf{x}}_h^d(t) + \dot\ell_i\mathbf{e}_i^d + \ell_i\dot{\mathbf{e}}_i^d,\\
    \ddot{\mathbf{x}}^{d}_i(t) &= \ddot{\mathbf{x}}_h^d(t) + \ddot\ell_i\mathbf{e}_i^d + 2\dot\ell_i\dot{\mathbf{e}}_i^d + \ell_i\ddot{\mathbf{e}}_i^d,
\end{align}

\subsection{Onboard Quadrotor Trajectory Tracking and Indirect Payload Control}\label{subsec:onboard_control}

\begin{figure}
    \centering
    \includegraphics[width=0.9\linewidth]{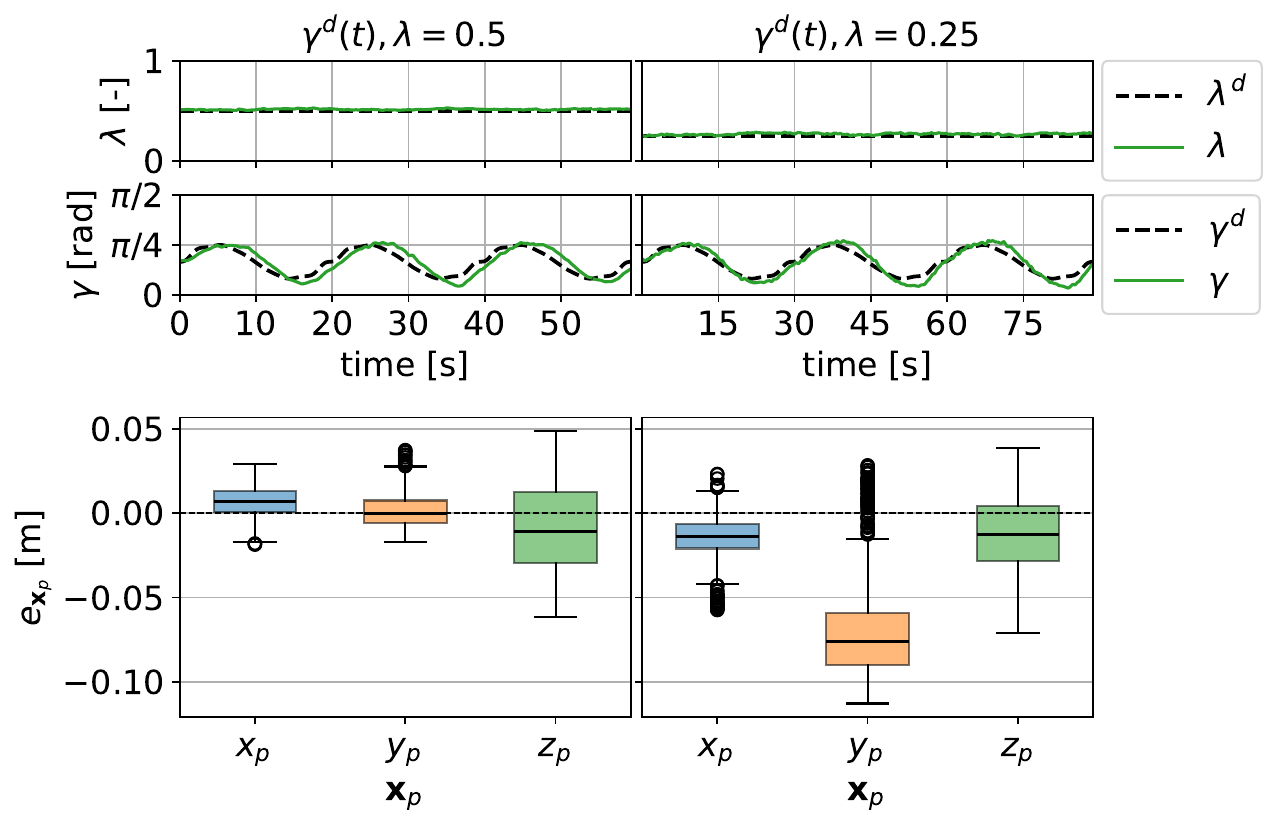}
    \caption{Experiment (i.b): Fixed payload position under $\gamma^d$ oscillations. Top: $\gamma^d(t)$ for two values of $\lambda^d$. Bottom: payload-position errors.}
    \label{fig:Ex1_gamma_swing}
\end{figure}

\begin{figure*}[t]
    \centering
    \includegraphics[width=0.9\linewidth, height=7cm]{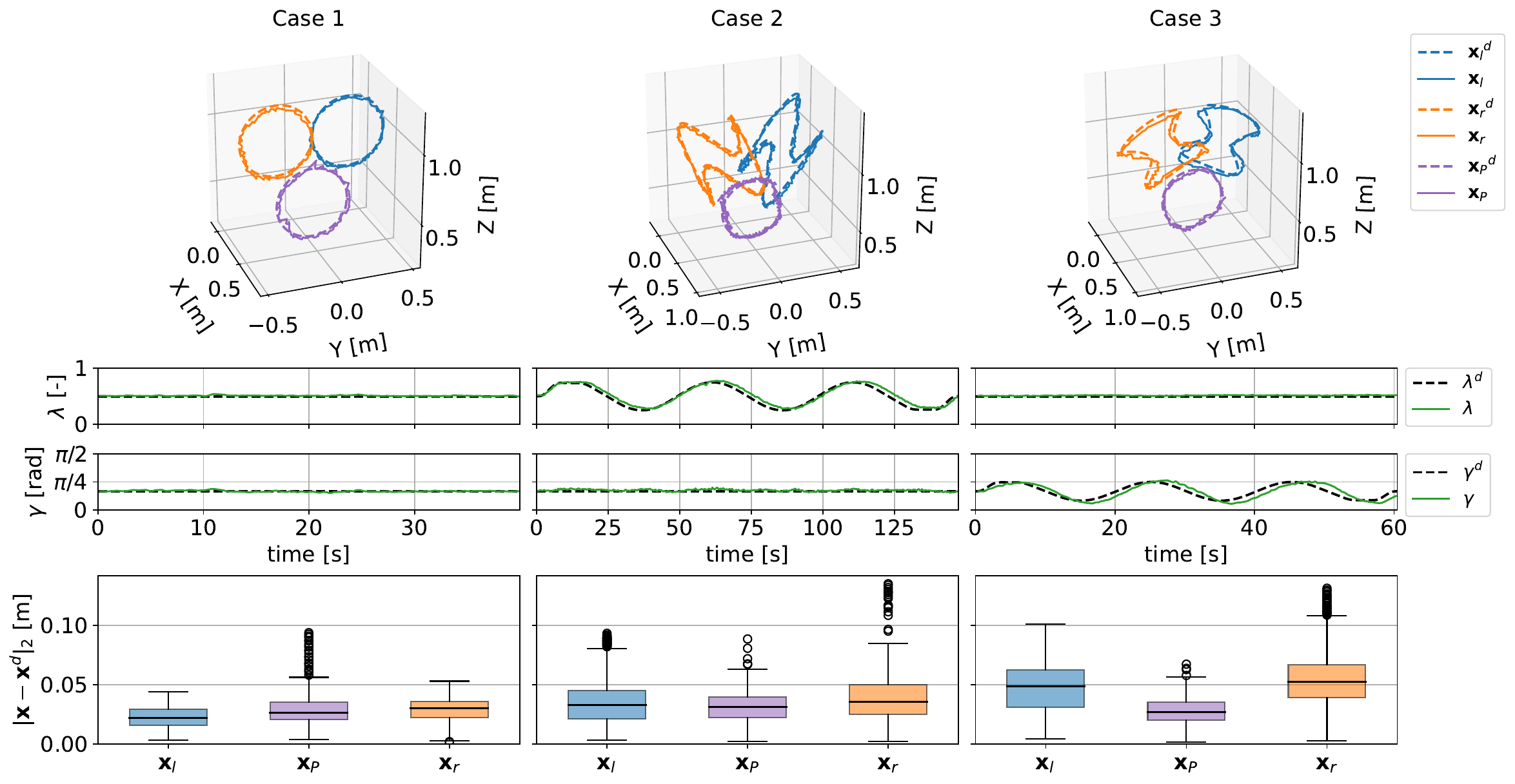}
    \caption{Experiment (ii). Payload path-following for a circular trajectory with appendant oscillations of $\lambda$ and $\gamma$ as disturbances. Going top to bottom, the first 3D-plot shows the planar trajectory reference and the measured position of the payload and quadrotors. The second one shows the time series of $\lambda^d(t)$ and $\gamma^d(t)$. And the third one is a whisker plot for the Euclidean position error for the left vehicle, the payload and the right vehicle.}
    \label{fig:Ex2_trajectory}
\end{figure*}

No additional payload-level feedback controller is implemented. Instead, payload position is controlled indirectly by commanding two quadrotors to track the geometrically reconstructed desired position \eqref{eq:drone-desired-position}, velocity, and acceleration \eqref{eq:drone-desired-vel-acc}.

Using onboard estimates of position, velocity, attitude, and angular velocity, the Crazyflie Mellinger controller generates the desired force \cite{Mellinger2011}
\begin{equation}
\begin{aligned}
    \mathbf{F}_i^d
    ={}& \mathsf{m}_i
    \left(\ddot{\mathbf{x}}_i^d+g\mathbf{e}_z\right)
    +K_{x_i}\left(\mathbf{x}_i^d-\hat{\mathbf{x}}_i\right) \\
    &+K_{v_i}\left(\dot{\mathbf{x}}_i^d-\hat{\dot{\mathbf{x}}}_i\right)
    +K_{I_i}\int_0^t
    \left(\mathbf{x}_i^d-\hat{\mathbf{x}}_i\right)d\zeta ,
\end{aligned}
\label{eq:onboard_desired_force}
\end{equation}
where $\mathsf{m}_i$ is the mass, and $K_{x_i}$, $K_{v_i}$,
and $K_{I_i}$ are the position, velocity, and integral gains.
The desired force direction and yaw define $R_i^d$,
while the collective thrust is
$f_i=\mathbf{F}_i^d\cdot R_i\mathbf{e}_z$.

The payload motion therefore results from the sliding cable coupling and the coordinated vehicles tracking, rather than direct payload feedback. The effects of cable tension, payload swing, and modeling errors enter the onboard controllers as unmodeled disturbances. Thus, successful tracking requires smooth references, a taut cable, accurate geometric reconstruction, and adequate trajectory by both quadrotors.

\section{Experiments}

DQ-HSTS was evaluated through three experiments: (i) maintaining a fixed payload position while varying $\lambda^d$ or $\gamma^d$; (ii) tracking a payload trajectory while both parameters oscillate; and (iii) transporting the payload beneath a ceiling obstacle. The platform comprises two Crazyflie 2.1 Brushless nano-quadrotors 
connected by a $1 m$ cable and localized by Lighthouse decks and three Lighthouse V2 base stations. References are generated offline and transmitted by a control computer.
Experiments use $\lambda\in[0.25,0.75]$, $\gamma\in[\pi/12,\pi/4]$ and the $V_{\mathrm{max}}=0.08,\mathrm{m/s}$; the remaining parameters are located within the repository's respective configuration files\footnote{\href{https://github.com/dmarslab/dh_psts_crazyflie_ws}{\texttt{https://github.com/dmarslab/dh\_psts\_crazyflie\_ws}}}.

A third tracked Crazyflie serves as the payload; its measurements are used only for validation, not feedback.

\subsection{Experiment i: Constant Payload Position
}
This experiment was implemented using 2 trajectories to understand $\lambda$'s and $\gamma$'s impact dependently on the payload hover performance. In Fig. \ref{fig:Ex1_lambda_swing} and \ref{fig:Ex1_gamma_swing} we compare the estimated $\lambda$ or $\gamma$ from the payload and vehicles position with their reference, and the error of payload position distribution. For each case, the desired waypoints $\mathbf{\xi}_\textrm{i}^d=\left[ \mathbf{x}_{p,\textrm{i}}^d, \lambda_\textrm{i}^d, \gamma_\textrm{i}^d \right]^\top$ where the payload reference was fixed on $\mathbf{x}_{p,\textrm{i}}^d(n) = [0.4, 0.0, 0.5]^\top$.

\paragraph{\textbf{varying $\lambda^d$}}\label{prg: exp1_a}
The trajectory was studied using two values of $\gamma_\textrm{i}^d(n) \in \left\{ \pi/12, \pi/4 \right\}$ while the the $\lambda^d$'s waypoints are:

\begin{equation}\label{eq:exp_ia}
    \lambda_\textrm{i}^d(n) = 0.25\sin{\left(2\pi n/N\right)} + 0.5, \qquad n=1,...N
\end{equation}


The experimental results are shown in Fig.~\ref{fig:Ex1_lambda_swing}, where the error estimates for $\lambda$ and $\gamma$ are $\textrm{RMSE}_\lambda=0.061$ and $\textrm{RMSE}_\gamma=0.027\, rad$. For the payload error, the mean error on x- and y-axis are very similar without care the $\gamma$ value, $\bar{x}_{p,\gamma=\pi/12}=0.9\,\text{cm}$, $\bar{x}_{p,\gamma=\pi/4}=1.0\,\text{cm}$, $\bar{y}_{p,\gamma=\pi/12}=0.5\,\text{cm}$ and $\bar{y}_{p,\gamma=\pi/4}=0.2\,\text{cm}$. However, for z-axis the values we can notice a important difference in the mean error from $\bar{z}_{p,\gamma=\pi/4}=0.3\,\text{cm}$ to $\bar{z}_{p,\gamma=\pi/4}=-5.7\,\text{cm}$; this change is expected, given that the magnitude of the cable $T \propto 1/\cos{\gamma}$, generating an additional overload in the quadrotors.

Finally, the most notable difference is the increase in the standard deviation of the error for the $y_p$ component, rising from $\sigma_{y_p,\gamma=\pi/12}=1.9\,\text{cm}$ to $\sigma_{y_p,\gamma=\pi/4}=5.6\,\text{cm}$. Although the model assumes
a frictionless hook, residual friction remains in the experimental setup. Cable sliding can therefore drag the hook along the cable, contributing to lateral tracking errors. This effect appears more pronounced at $\gamma=\pi/4$, where the cable segments are more horizontal.

\paragraph{\textbf{varying $\gamma^d$}}\label{prg: exp1_b}

In this case, the trajectory was evaluated under two values of $\lambda_\textrm{i}^d(n) \in \left\{ 0.5, 0.25 \right\}$ while the $\gamma^d$'s waypoints are:
\begin{equation}\label{eq:exp_ib}
    \gamma_\textrm{i}^d(n) = \pi/12\sin{\left(2\pi n/N\right)} + \pi/6 \qquad n=1,...N
\end{equation}

Fig. \ref{fig:Ex1_gamma_swing} shows the estimation of $\lambda$ and $\gamma$, with errors of $\textrm{RMSE}_\lambda = 0.019$ and $\textrm{RMSE}_\gamma = 0.105\, \text{rad}$. The mean error in the load position is practically the same for the $z$-axis: $\bar{z}_{p,\lambda=0.5}=-0.8\,\text{cm}$ and $\bar{z}_{p,\lambda=0.25}=-1.2\,\text{cm}$; in contrast, for $x_p$ it varies from $\bar{x}_{p,\lambda=0.5}=0.6\,\text{cm}$ to $\bar{x}_{p,\lambda=0.25}=-1.4\,\text{cm}$, while for $y_p$ the change is more pronounced, shifting from $-0.2\,\text{cm}$ to $-7.0\,\text{cm}$.

The asymmetric configuration ($\lambda=0.25$) tilts the spheroid's focal axis toward the vertical, bringing the hook closer to the lower quadrotor. This asymmetric configuration can increase sensitivity to disturbances and promote horizontal oscillation of the load. Given that the cable reaction forces act primarily in the $\Pi_{y,z}$ plane, their lateral effects may contribute more significantly to the error in $y_p$ than to the error in $x_p$, which is consistent with the greater bias observed in $y_p$.

\subsection{Experiment ii: Tracking a payload trajectory while $\lambda$ and $\gamma$
oscillate}

For this test, we generate a circular trajectory for the payload around $|Pi_{y,z}$ plane, while the parameters $\lambda^d$ and $\gamma^d$ oscillates as was shown in the experiments \eqref{eq:exp_ia} and \eqref{eq:exp_ib}. The desired payload waypoints are

\begin{equation}\label{eq:exp_ii}
    \mathbf{x}_{p,\textrm{ii}}^d(n) = 
    \begin{bmatrix}
        0.4\\
        0.25 \cos{\left(2\pi n/N\right)}\\
        0.25 \sin{\left(2\pi n/N\right)}+0.5
    \end{bmatrix} \qquad n=1,...N,
\end{equation}
and the experiment was evaluated in 3 cases, 
\begin{itemize}
    \item \textbf{Case 1}: $\lambda_\textrm{ii}^d = 0.5$ and $\gamma_\textrm{ii}^d = \pi/6$
    \item \textbf{Case 2}: $\lambda_\textrm{ii}^d(n) = \lambda_\textrm{i.a}^d(3n)$ and $\gamma_\textrm{ii}^d = \pi/6$
    \item \textbf{Case 3}: $\lambda_\textrm{ii}^d = 0.5$ and $\gamma_\textrm{ii}^d(n) = \gamma_\textrm{i.b}^d(3n)$
\end{itemize}
The Fig.~\ref{fig:Ex2_trajectory} confirm some things that the previews experiment shows, the bigger estimation error of $\lambda$ and $\gamma$ was $\textrm{RMSE}_\lambda=0.05$ in case 2 and $\textrm{RMSE}_\gamma=0.11\, \text{rad}$ in case 3. However, the payload trajectory tracking error are very similar for the 3 cases, around $\textrm{RMSE}_{x_p}=3\, \text{cm}$. The bigger mean euclidean error for the payload tracking was in case 2 with $\overline{x_p}=3.2\, \text{cm}$. However, this test is very important because it shows us that the proposed indirect control strategy is effective enough to make the load follow a trajectory subject to the $V_{\mathrm{max}}$ constraint.

\subsection{Experiment iii: Transporting the payload beneath a ceiling obstacle}
\begin{figure}
    \centering
    \begin{subfigure}[b]{0.48\textwidth}
        \centering
        \includegraphics[width=1\linewidth]{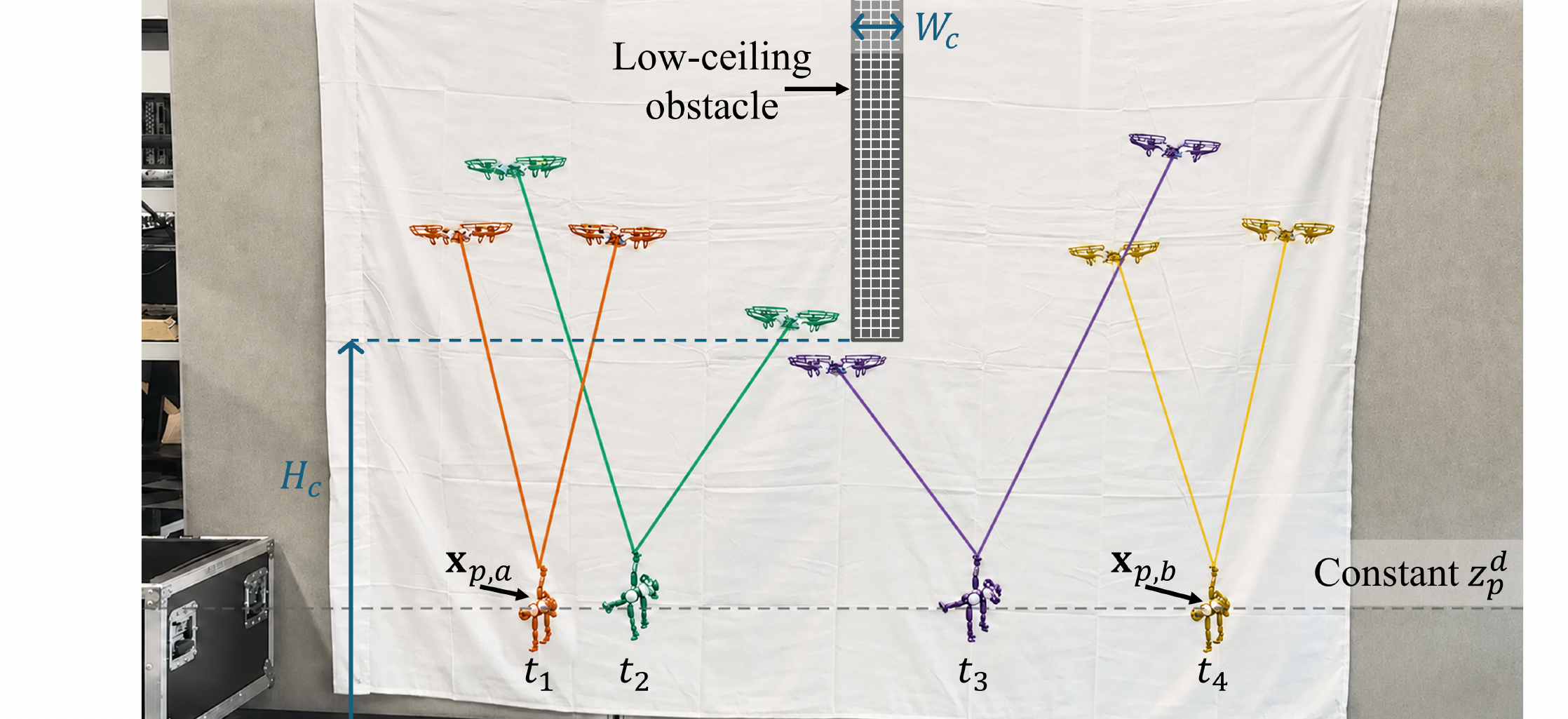}
        \caption{}
        \label{fig:Ex3_obstacle_a}
    \end{subfigure}
    \hfill
    \begin{subfigure}[b]{0.48\textwidth}
        \centering
        \includegraphics[width=1\linewidth]{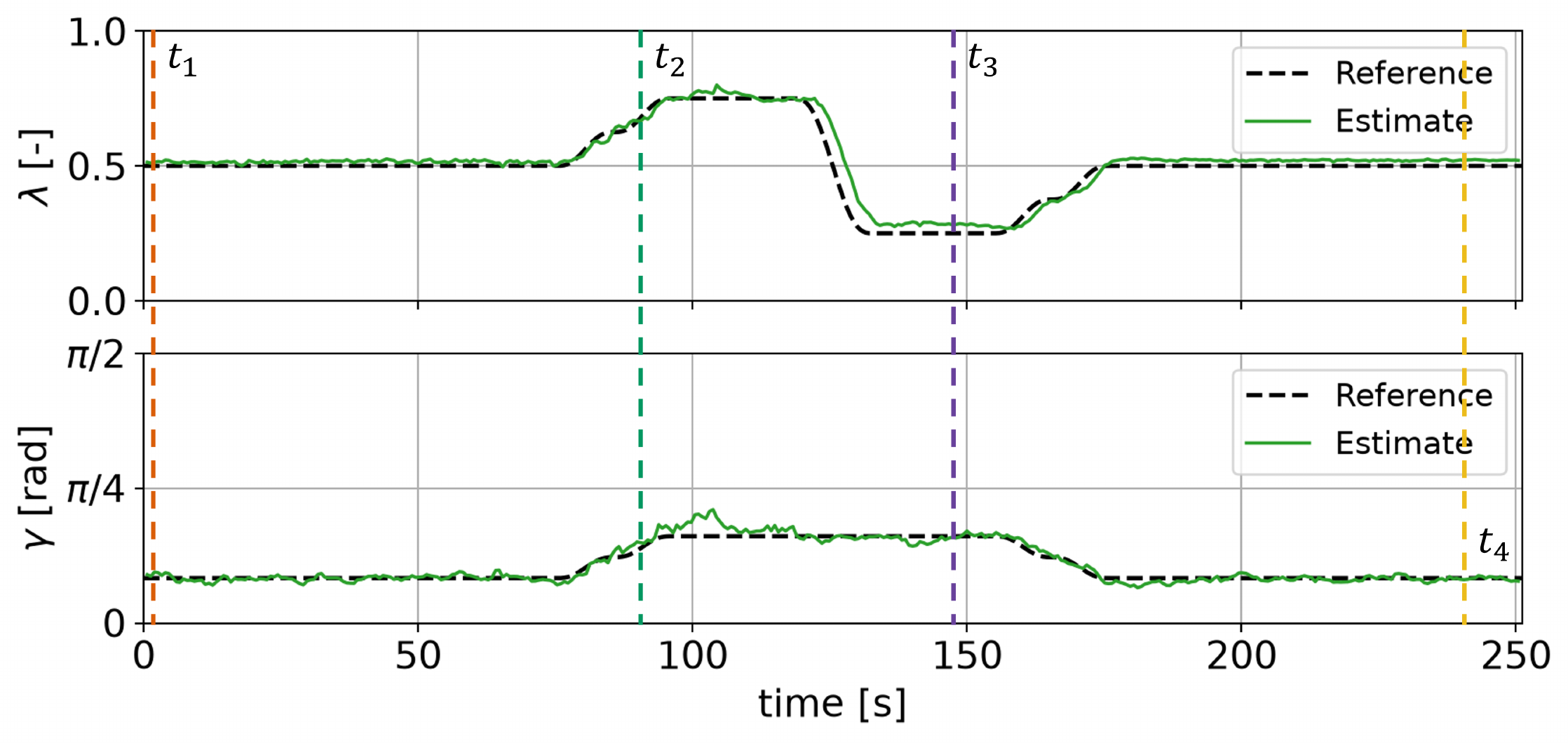}
        \caption{}
        \label{fig:Ex3_obstacle_b}
    \end{subfigure}
    \caption{Experiment (iii). (a) Time lapse of the DQ-HSTS robot moving the payload horizontally from right to left avoiding a low-ceiling obstacle. (b) Time evolution of $\lambda$ and $\gamma$ during the mission.}
    \label{fig:Ex3_obstacle}
    \vspace{-1.5em}
\end{figure}

This experiment aims to demonstrate the ability of DQ-HSTS to navigate around potential obstacles, highlighting the advantages of our robot's additional degree of freedom. To this end, we designed and implemented a mission for our robot that highlighted the advantage of the sliding hook over a fixed hitch:

\begin{mission}
Transport the payload horizontally from $\mathbf{x}_{p,a}$ to $\mathbf{x}_{p,b}$ beneath a low-ceiling obstacle of height $H_c$, width $W_c$, and center $\mathbf{x}_c$ midway along the path.
\end{mission}

To achieve this, a pre-designed trajectory varied $\lambda$ and $\gamma$ to pass beneath the obstacle while limiting the tension increase associated with increasing $\gamma$ alone. Fig.~\ref{fig:Ex3_obstacle_a} shows the motion, with a mean payload Euclidean tracking error of $3.19\,\text{cm}$, a standard deviation of $1.32\,\text{cm}$, and an average speed of $0.038\,\text{m/s}$. Fig.~\ref{fig:Ex3_obstacle_b} compares the reference and estimated configuration parameters, yielding $\mathrm{RMSE}_{\lambda}=0.03$ and $\mathrm{RMSE}_{\gamma}=0.03\,\text{rad}$.

\section{Conclusions}

This work presented, analyzed, and experimentally validated the DQ-HSTS, a novel transport architecture consisting of two quadrotors connected by a single cable that slides through an ideally frictionless hook attached to the payload. Unlike conventional multi-quadrotor transport systems, in which each cable is directly attached to the payload, the proposed architecture introduces the cable-length distribution $\lambda$ as an additional configuration degree of freedom. This enables the cable-segment lengths to vary while preserving the desired payload position.

The experimental results showed that the proposed geometric model enables accurate indirect tracking of the payload position despite this additional degree of freedom. Nevertheless, tracking performance was sensitive to particular combinations of $\lambda$ and $\gamma$ and deteriorated as $\gamma$ increased. Under quasi-static equilibrium, the required cable tension is proportional to $1/\cos\gamma$ and therefore increases with $\gamma$. This theoretical behavior is consistent with the experimentally observed reduction in the mean payload height $z_p$ at larger values of $\gamma$.


In the third experiment, $\lambda$ and $\gamma$ were pre-calculated to avoid a known low-ceiling obstacle while maintaining a constant payload trajectory. This experiment took advantage the flexibility of the DQ-HSTS and kept the Euclidean error of the payload position within the range demonstrated in previous experiments. Eliminating the exclusive dependency on $\gamma$ for obstacle avoidance, given its correlation with the load the drones must carry, shows the potential of $\lambda$ for transport operations that require obstacle avoidance without excessively increasing the load on the vehicles; this, in turn, motivates the future development of automated methods for obstacle-avoiding trajectory generation and configuration optimization.

Overall, the DQ-HSTS achieved effective payload transportation through geometric reconstruction and indirect position control using the quadrotor references. Nevertheless, direct closed-loop payload control was beyond the scope of this study. Future work will incorporate payload-state feedback to improve robustness against cable dynamics, payload swing, quadrotor tracking errors, and modeling uncertainties.

\bibliographystyle{ieeetr}
\bibliography{references.bib}\

\clearpage

\end{document}